\documentclass[conference]{IEEEtran}
\IEEEoverridecommandlockouts

\usepackage{cite}
\usepackage{hyperref}
\usepackage{amsmath,amssymb,amsfonts}
\usepackage{graphicx}
\usepackage{textcomp}
\usepackage{xcolor}
\usepackage{booktabs}
\usepackage{array}
\usepackage{url}
\usepackage{microtype}

\def\BibTeX{{\rm B\kern-.05em{\sc i\kern-.025em b}\kern-.08em
    T\kern-.1667em\lower.7ex\hbox{E}\kern-.125emX}}

\begin{document}

\title{Evidence-Gated Bilingual RAG for Reliable Consumer Banking Search}

\author{
\IEEEauthorblockN{Ahmed Mohamed Mahmoud Rayyan}
\IEEEauthorblockA{\textit{School of Computing}\\
\textit{Queen's University}\\
Kingston, Canada\\
hamadarayyan056@gmail.com}
\and
\IEEEauthorblockN{Aya Abdallah Abdelmonem}
\IEEEauthorblockA{\textit{School of Computing}\\
\textit{Queen's University}\\
Kingston, Canada\\
ayaabdallaha1010@gmail.com}
\and
\IEEEauthorblockN{Mohamed Khaled Goda Othman}
\IEEEauthorblockA{\textit{School of Computing}\\
\textit{Queen's University}\\
Kingston, Canada\\
Engmohamedsw@gmail.com}
\and
\IEEEauthorblockN{Rana Mahmoud Mohammed}
\IEEEauthorblockA{\textit{School of Computing}\\
\textit{Queen's University}\\
Kingston, Canada\\
ranaa208200@gmail.com}
\and
\IEEEauthorblockN{Rahatara Ferdousi}
\IEEEauthorblockA{\textit{School of Computing}\\
\textit{Queen's University}\\
Kingston, Canada\\
rahatara.ferdousi@queensu.ca}
}

\maketitle

\begin{abstract}
Consumers often describe what they need rather than use the terminology found on a banking website. For example, a natural-language request (``a card for travel'') may not share the keywords of the relevant product page, and the problem is harder in bilingual services, where the same intent may be expressed in Arabic or English. This paper presents an evidence-gated bilingual retrieval-augmented generation (RAG) framework for consumer banking search that operates entirely on local infrastructure. The framework combines structured web-content refinement, hybrid dense--lexical retrieval, intent-aware evidence selection, cross-encoder reranking, and a three-outcome Answer / Related Mismatch / No Answer decision mechanism. On 210 URL-labeled Arabic and English queries, the proposed approach improves Top-1 hit rate from 50.48\% to 79.05\% and MRR@5 from 0.571 to 0.797 over a hybrid baseline. A separate 26-case end-to-end evaluation achieved 69.2\% strict success, with 72.7\% of generated answers classified as supported by an automated faithfulness judge. These results are preliminary: the retrieval queries are paraphrases of a small reference set, and the end-to-end test is small and uses an automated rather than human judge, so they do not establish human-verified answer correctness or generalizable performance.
\end{abstract}


\section{Introduction}

A consumer may search a bank's website for ``a card for travel,'' while the relevant information is organized under specific credit-card products and benefits. This vocabulary mismatch limits conventional keyword-based search, particularly when consumers express their needs in natural language. In multilingual markets such as Egypt, the challenge is amplified because the same need may be expressed in Arabic or English using terminology that differs from the published product pages. This is particularly relevant to consumer technology, where language-model-based services have received increasing industry attention \cite{scott2025}.

\begin{figure*}[t]
\centering
\includegraphics[width=0.8\textwidth]{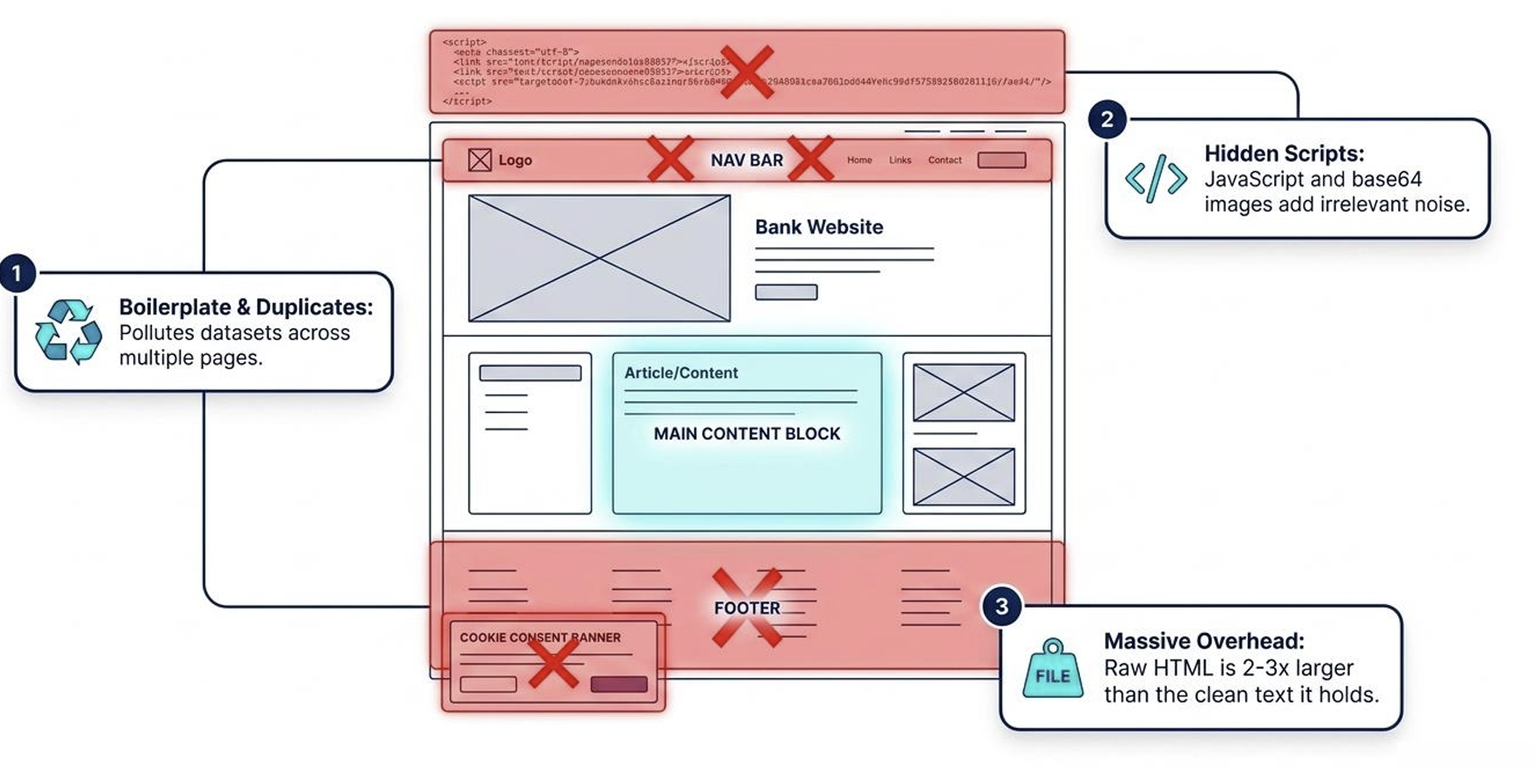}
\caption{Sources of retrieval-irrelevant content in raw consumer banking webpages prior to retrieval-oriented content refinement.}
\label{fig:noise}
\end{figure*}

Improving search requires more than replacing keyword matching with semantic retrieval. Consumer banking websites are not necessarily organized as retrieval-ready knowledge bases. As illustrated in Fig.~\ref{fig:noise}, raw webpages contain navigation elements, repeated boilerplate, scripts, embedded media, and other content that can obscure information relevant to retrieval. The studied National Bank of Egypt (NBE) website also contains dynamically rendered content, motivating a structured extraction and refinement pipeline.

Dense semantic retrieval can reduce vocabulary mismatch \cite{karpukhin2020}, while combining semantic and lexical retrieval can improve coverage across different query formulations \cite{ma2021}. However, semantic relevance alone does not establish that retrieved content belongs to the intended banking service or provides sufficient evidence for an answer. This distinction is important in consumer financial applications, where plausible but mismatched evidence can lead to unsupported responses.

To address these challenges, we propose an \emph{evidence-gated bilingual retrieval-augmented generation (RAG) framework} \cite{lewis2020} for consumer banking search. The framework transforms raw web content into structured, retrieval-ready evidence and combines Arabic--English hybrid retrieval with intent-aware evidence selection and an explicit Answer / Related Mismatch / No Answer decision layer. This layer separates finding semantically relevant content from determining whether the retrieved evidence is sufficient and consistent with the intended banking service. The complete framework operates on local infrastructure, avoiding the need to send consumer queries to an external cloud AI service.

This paper makes three contributions:
\begin{enumerate}
    \item \textbf{A retrieval-ready data refinery for banking web content.} We transform noisy, dynamically rendered bilingual banking webpages into structured, deduplicated evidence while preserving provenance and product information.
    
    \item \textbf{An evidence-gated bilingual RAG framework.} We combine hybrid dense--lexical retrieval with intent-aware evidence selection and explicit answer eligibility to distinguish sufficient evidence from mismatched or insufficient evidence.
    
    \item \textbf{A bilingual consumer-banking evaluation.} We evaluate retrieval and evidence eligibility on 220 Arabic and English queries, with ranking metrics reported on the 210 URL-labeled queries, where the proposed framework improves Top-1 hit rate from 50.48\% to 79.05\% over the hybrid baseline. We also report a separate, small 26-case end-to-end evaluation of generation, citation, and abstention behavior.
\end{enumerate}

The source code for the project is available on GitHub.\footnote{\url{https://github.com/DigiGradProject/SmartBankSearch}}

\section{Related Work}
\label{rw}
\subsection{Hybrid and Multilingual Retrieval}
Retrieval-augmented generation grounds language-model responses in external evidence, making retrieval quality central to downstream reliability \cite{lewis2020}. Sparse methods such as BM25 remain effective when queries share explicit terminology with relevant documents \cite{robertson2009}, while dense retrieval can capture semantic similarity beyond exact lexical overlap \cite{karpukhin2020}. Hybrid retrieval combines these complementary signals \cite{ma2021}, and cross-encoder reranking can further refine candidate ordering \cite{nogueira2019}. Recent financial RAG systems similarly report benefits from combining full-text and semantic retrieval and from neural reranking over domain-specific documents \cite{cheng2026}.

Multilingual retrieval introduces additional challenges because equivalent information needs may be expressed through different linguistic forms. BGE-M3 supports dense and sparse retrieval across more than 100 languages and provides a common representation for multilingual retrieval \cite{chen2024}. Recent multilingual RAG studies nevertheless report persistent issues, including irrelevant retrieval, code-switching, and language-specific generation errors \cite{chirkova2024}. Arabic RAG presents additional variation between query and document language forms, including differences in dialect and expression \cite{elbeltagy2024}, and consumer-facing Arabic systems must also cope with colloquial and code-switched formulations \cite{hamed2025} and with orthographic variation \cite{obeid2020}. These challenges are particularly relevant to consumer banking search, where users describe a need rather than reproduce the formal terminology of a financial product.

\subsection{Reliability and Evidence-Aware RAG}
Retrieving semantically related content does not guarantee that the evidence is sufficient to support an answer. Recent RAG research therefore increasingly treats retrieval quality as part of the generation decision. Self-RAG introduces adaptive retrieval and self-reflection to assess retrieved passages and generated responses \cite{asai2024}. Corrective RAG evaluates retrieved documents and triggers corrective retrieval when evidence is inaccurate or ambiguous \cite{yan2024}. These approaches demonstrate the importance of evaluating evidence rather than automatically conditioning generation on retrieved passages.

Abstention provides another reliability mechanism when available evidence cannot support a response. Recent evaluations show that deciding when not to answer remains difficult even for modern reasoning models \cite{kirichenko2025}, which motivates explicit mechanisms for handling insufficient evidence and mitigating hallucination \cite{ji2023}. In consumer financial applications, a semantically related result may belong to the wrong product or service category and should not automatically become evidence for an answer. Reliability of consumer-facing chatbots is likewise a concern in the consumer-electronics literature, for example with respect to adversarial attacks on vision-language-model-empowered chatbots \cite{shang2024}.

\subsection{Consumer Banking RAG and Research Gap}
Financial RAG research has primarily examined retrieval and question answering over financial reports and other document collections, where hybrid retrieval, domain adaptation, and reranking have shown benefits \cite{cheng2026}. Consumer banking search is a different setting: users formulate short natural-language information needs, relevant evidence is distributed across product and service pages, and retrieved content must correspond to the intended banking service before it can safely support generation. Within the consumer-electronics community, language-model services are increasingly studied for consumer and edge devices, including privacy-oriented federated and edge--cloud LLM frameworks \cite{zhang2025,otoum2026}, LLM-based audio processing for consumer devices \cite{huang2026}, and consumer-facing dialogue systems such as multi-turn medical dialogue generation \cite{wang2025}. These studies do not address search over consumer banking content, but they show the growing role of language models in consumer-technology settings and motivate local, resource-conscious deployment. Locally deployed toolkits show that complete RAG pipelines can operate without external cloud APIs \cite{maiya2025}, but they do not address this eligibility question. Our work addresses this \emph{evidence-eligibility gap} by gating generation on evidence sufficiency rather than retrieval relevance alone.

\begin{figure*}[t]
\centering
\includegraphics[width=\textwidth]{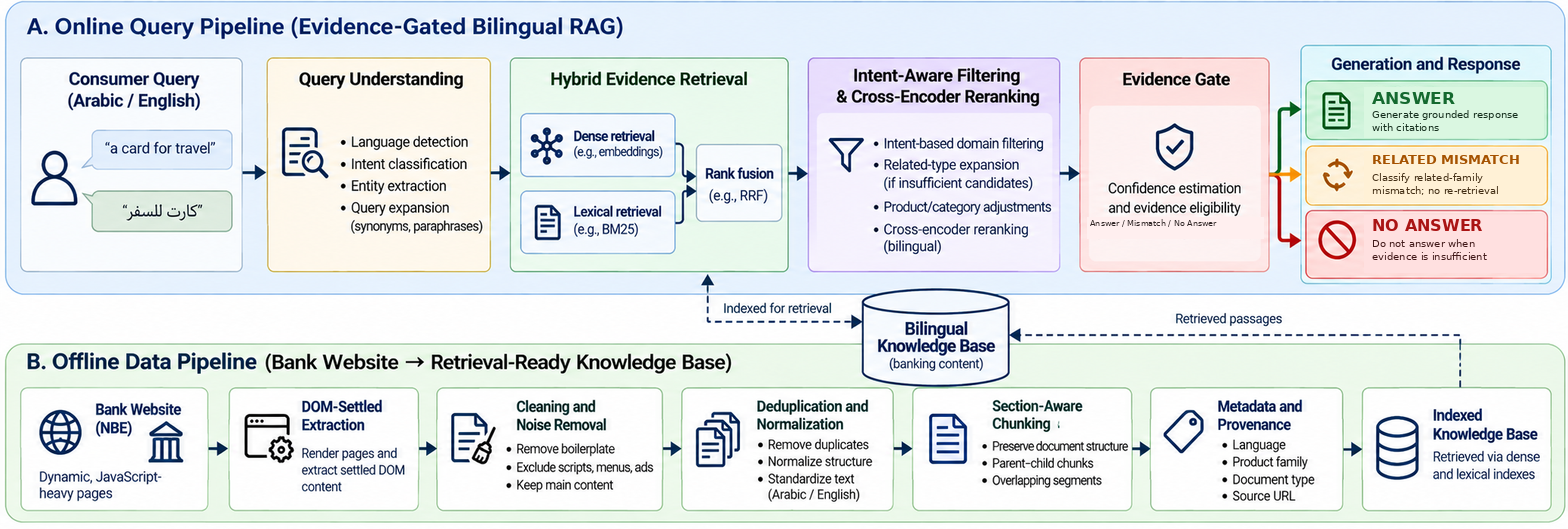}
\caption{Architecture of the proposed evidence-gated bilingual RAG framework. The offline pipeline builds a structured bilingual evidence base; the online pipeline resolves each query to one of three outcomes: Answer, Related Mismatch (a classification label, not a second retrieval), or No Answer. Grounded generation and faithfulness checking are evaluated in a separate, preliminary 26-case end-to-end experiment reported in Section~\ref{sec:e2e}; they are not evaluated in the 220-query retrieval experiment. All components run on local infrastructure.}
\label{fig:arch}
\end{figure*}

\section{Proposed Framework}
\label{proposed}
Our proposed framework is illustrated in
Fig.~\ref{fig:arch}. An offline phase refines raw web content into structured evidence and indexes it; an online phase processes each query through understanding, hybrid retrieval, intent-aware evidence selection, and a decision layer that either releases the evidence for grounded generation or returns an explicit outcome without an answer. All components run on local infrastructure.

\subsection{Data Refinery and Knowledge Base Construction}
\begin{figure*}[t]
\centering
\includegraphics[width=\textwidth]{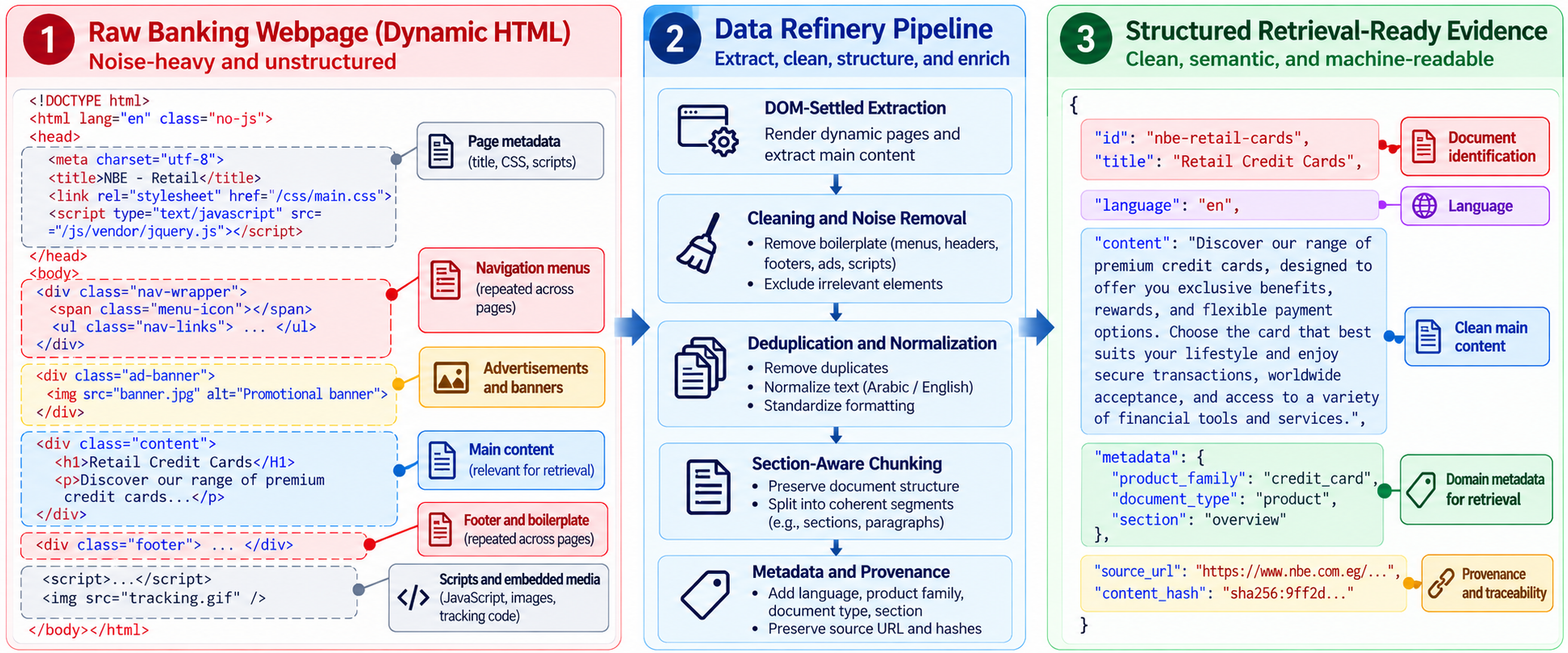}
\caption{Transformation of raw banking web content into structured, retrieval-ready evidence through the data refinery pipeline.}
\label{fig:refinery}
\end{figure*}

The data refinery (Fig.~\ref{fig:refinery}) transforms raw banking webpages into retrieval-ready evidence. Pages whose static content is thin because of dynamic client-side rendering are detected and re-acquired through browser-based rendering. Statistical boilerplate removal strips navigation, headers, footers, and repeated templates; Arabic text is orthographically normalized; tables are preserved as inline text; and near-duplicate pages are filtered. Each cleaned document follows a fixed contract (identifier, title, source URL, language, content, and metadata) and is split into section-aware parent and child chunks. Parent sections use an approximate 800-token budget, and indexed child windows use up to 280 tokens with a 50-token overlap. Records retain document identifiers, source URLs, language, product family, document type, and section information. Content hashes make re-ingestion idempotent, so unchanged chunks are not re-processed. The resulting knowledge base contains 555 effective documents and 962 chunks in Arabic and English.

\subsection{Dense and Lexical Indexing}
A multilingual embedding model, BGE-M3 \cite{chen2024}, produces dense vectors and sparse lexical weights for each chunk. Dense vectors and metadata are stored in a cosine-distance vector index, and a parallel BM25 index \cite{robertson2009} stores tokenized titles and text, with Arabic or Latin tokenization selected according to the document language.

\subsection{Query Understanding}
Each query passes through language detection, Arabic normalization, intent classification, entity extraction, query expansion, and rewriting. Deterministic rules for contact and authentication queries precede semantic intent classification. Embedding-based intent prototypes provide the primary semantic signal, with pattern-based fallbacks and a general-query route when evidence is weak. A query planner can split a question expressing distinct intents into at most two subqueries and merge their retrieved evidence; the retrieval evaluation in Section~\ref{exp} measures the retrieval stage alone rather than this complete path.

\subsection{Hybrid Retrieval and Reranking}
Candidate retrieval combines two ranked channels. The dense channel obtains up to 50 cosine-nearest candidates and rescores them using dense similarity and sparse overlap, with a default dense weight of 0.7. The lexical channel retrieves up to 50 positive-scoring BM25 matches. Reciprocal Rank Fusion (RRF) \cite{cormack2009} combines the two ranked lists:
\begin{equation}
\mathrm{RRF}(d)=\sum_{c\in\{\text{dense},\,\text{BM25}\}}\frac{1}{k+r_c(d)},
\label{eq:rrf}
\end{equation}
where $r_c(d)$ is the rank of document $d$ in channel $c$ (ranks start at one) and $k=60$. Fused scores are rescaled and capped at one, and up to 32 candidates proceed. Low-value and navigation-heavy material is filtered and keyword relevance adjusts the ordering. A cross-encoder reranker (BAAI/bge-reranker-v2-m3) \cite{nogueira2019,bge_reranker_v2_m3} then reranks a pool of up to 20 candidates and retains five. The BGE model card identifies this checkpoint as a multilingual reranker.

\subsection{Intent-Aware Evidence Selection}
In the full framework, intent-conditioned metadata filtering narrows the initial candidate pool to documents matching the classified banking intent. When filtered retrieval yields too few candidates, the retriever expands first to related document types and, if necessary, to broader language-restricted retrieval. Category, metadata, and product-family adjustments operate before and after reranking, reordering retrieved evidence without injecting documents absent from retrieval.

\subsection{Answer / Related Mismatch / No Answer Decision Layer}
The retrieval-only experiment evaluates the decision layer's eligibility outcomes from retrieval confidence and evidence-family matching; it does not invoke language-model generation. In the deployed end-to-end path, an eligible result can be passed to generation. The three decision outcomes are:
\begin{itemize}
\item \textbf{Answer:} evidence is present, intent confidence permits the gate, and the leading result matches the classified banking family through document type, URL markers, or category. In production, a compressed context of up to 2{,}000 tokens with up to three cited sources is passed to a locally hosted Qwen3 model \cite{qwen2025} for grounded generation.
\item \textbf{Related Mismatch:} when intent confidence is at least 0.90 but the leading result does not match the expected document family, the system records a related-family mismatch and returns an explicit outcome without an answer. This is a classification label, not a retrieval command: no second retrieval is launched after this decision, and any candidate broadening to related document types or a wider language-restricted search occurs earlier, inside the retrieval stage.
\item \textbf{No Answer:} evidence is absent, confidence falls below a threshold of 0.42, or the leading result does not match the expected document family under the remaining checks. An explicit message with an auditable reason is returned in place of a generated answer.
\end{itemize}
In the full end-to-end path, a faithfulness check evaluates whether factual claims in a generated answer are supported by the retrieved context. An unsupported answer triggers one stricter regeneration; if the regenerated answer remains unsupported, the system abstains. This generation-and-checking path is assessed separately in the 26-case experiment in Section~\ref{sec:e2e}, not in the 220-query retrieval experiment.

\subsection{Confidence Estimation}
Confidence is an engineered retrieval-evidence score, not a calibrated probability of answer correctness. Its core combines six signals: intent confidence (0.15), leading ranking score (0.25), mean of the top five scores (0.15), category consistency (0.20), score agreement (0.15), and source diversity (0.10). A high-confidence intent--metadata disagreement scales the core by 0.55, and the core is then blended at weight 0.75 with a keyword-based relevance score at weight 0.25. A language-model sufficiency component exists in the full framework but was inactive in the 220-query retrieval experiment.

\section{Experimental Setup}
\label{exp}
\subsection{Dataset and Knowledge Base}

The corpus was constructed from publicly available National Bank of Egypt (NBE) webpages using browser-based rendering to capture dynamically loaded content. After content refinement, boilerplate removal, and near-duplicate filtering, the bilingual knowledge base contains 555 documents and 962 Arabic and English chunks.

\subsection{Evaluation Set}

Retrieval is evaluated on 220 bilingual queries comprising 127 Arabic and 93 English questions across 15 banking intent categories. The queries are paraphrases of a 22-query reference set with no exact question overlap. Of the 220 queries, 210 have expected URLs and are used for ranking evaluation, while all 220 are used to evaluate evidence-eligibility decisions and forbidden-URL behavior. The set includes 43 queries with specified forbidden URLs, and exchange-rate and personal-loan queries account for 142 queries (64.55\%). Arabic--English code-switched queries are not included. As the reference set also informed some domain rules, the results characterize performance in the present test setting rather than a fully independent benchmark.

\subsection{Retrieval Configurations}

We compare two retrieval configurations. \emph{Hybrid Baseline} combines BM25, sparse rescoring, query expansion, keyword ranking, and cross-encoder reranking. \emph{Evidence-Gated} extends this pipeline with intent-aware evidence selection, domain-aware ranking, and confidence-based Answer / Related Mismatch / No Answer decisions. The comparison therefore evaluates the complete proposed configuration against a hybrid retrieval baseline; because several components change together, it does not isolate the contribution of any single component.

\subsection{Models and Infrastructure}

BGE-M3 is used for multilingual embedding and BGE-Reranker-v2-M3 for cross-encoder reranking. Qwen3-8B \cite{qwen2025} is served locally through Ollama and is used only in the separate end-to-end evaluation, both for response generation and for automated faithfulness assessment; the 220-query experiment evaluates retrieval and evidence eligibility independently of generation. Retrieval experiments were conducted on a single 16~GB GPU. In the 26-case end-to-end evaluation, generated answers were automatically classified as supported, partially supported, or unsupported based on the retrieved evidence.

\subsection{Evaluation Metrics}

Retrieval performance is evaluated using Top-1, Top-3, Hit@5, MRR@5, and nDCG@5 on the 210 URL-labeled queries. Top-$k$ measures whether the expected URL appears within the first $k$ retrieved results (Hit@5 is therefore a hit rate, not recall over all relevant documents), while MRR@5 and nDCG@5 measure ranking quality within the top five results. Evidence-eligibility outcomes and forbidden-URL violations are evaluated over all 220 queries. The separate end-to-end evaluation reports strict success, abstention accuracy, source retrieval and citation rates, faithfulness, and response latency. Strict end-to-end success requires the expected answer or abstain behavior, a supported answer where an answer is expected, and correct source citation.

\section{Results and Discussion}
\label{res}

\subsection{Retrieval Performance}

Table~\ref{tab:retrieval} compares the proposed Evidence-Gated framework with the hybrid baseline on the 210 URL-labeled queries. The proposed framework improves Top-1 hit rate from 50.48\% to 79.05\%, a gain of 28.57 percentage points. Improvements are also observed for Top-3, Hit@5, MRR@5, and nDCG@5, indicating that the proposed framework ranks expected banking pages higher on this dataset. The largest gain occurs at rank 1, which is particularly relevant to consumer-facing search, where the first result strongly influences information access. Results come from a single run per configuration, without significance testing, and because several components differ between the two configurations, the effect of any one component is not isolated.

\begin{table}[t]
\caption{Retrieval Performance on 210 URL-Labeled Queries}
\label{tab:retrieval}
\centering
\footnotesize
\setlength{\tabcolsep}{4pt}
\begin{tabular}{lccc}
\toprule
Metric & Hybrid & Proposed & Improvement \\
\midrule
Top-1 (\%)  & 50.48 & 79.05 & +28.57 \\
Top-3 (\%)  & 62.38 & 79.52 & +17.14 \\
Hit@5 (\%)  & 70.00 & 81.43 & +11.43 \\
MRR@5       & 0.5708 & 0.7974 & +0.2266 \\
nDCG@5      & 0.5950 & 0.7975 & +0.2025 \\
\bottomrule
\end{tabular}
\end{table}

\subsection{Evidence Eligibility and Efficiency}

Beyond ranking relevant pages, the proposed framework evaluates whether retrieved evidence is sufficient for answering. It classified 214 of 220 queries (97.3\%) as answerable and abstained on six (2.7\%) under its decision criteria, whereas the hybrid baseline returned results for all queries. No forbidden-URL violations were observed for the proposed framework, compared with one for the baseline. The Related Mismatch outcome was not triggered in this evaluation. These outcomes describe retrieval eligibility; the correctness of the abstentions was not independently assessed. Mean retrieval latency was 2.82~s for the proposed framework and 3.07~s for the baseline (95th percentile 3.44~s versus 3.33~s); these are single-run, retrieval-only, environment-specific figures from a shared GPU, not a controlled benchmark.

\subsection{End-to-End Evaluation}
\label{sec:e2e}

Table~\ref{tab:e2e} summarizes a separate end-to-end run of 26 cases: the 22-query retrieval reference set plus four explicit abstention cases. The framework achieved 69.2\% strict end-to-end success (18/26) and correctly abstained on all four designated abstention cases. The expected source was retrieved in all 17 cases with annotated gold sources, and 15 of these cases (88.2\%) included the expected source in the generated citations. Expected-source URL match was 69.4\% micro-averaged (25 of 36 emitted citations across those 17 cases); this is a URL-pattern measure, not claim-level citation precision. No forbidden-source citations were observed in the five cases with a specified forbidden source. Among the 22 generated answers, an automated Qwen3 faithfulness judge classified 72.7\% as supported, 22.7\% as partially supported, and 4.5\% as unsupported. Mean end-to-end latency was 17.07~s on the local deployment.

These results are preliminary. The 22 answerable cases are the same reference set that informed several domain rules, so the run is not an independent test of generalization, and the 100\% gold-source retrieval rate should be read with this overlap in mind; it is higher than the Hit@5 of 81.43\% on the 210-query set. The judge is the same model family used for generation (Qwen3-8B), is not human adjudication, and URL matching does not verify individual factual claims.

\begin{table}[t]
\caption{End-to-End Evaluation on 26 Cases}
\label{tab:e2e}
\centering
\footnotesize
\setlength{\tabcolsep}{3pt}
\begin{tabular}{lc}
\toprule
Metric & Result \\
\midrule
Answerability accuracy              & 100.0\% (26/26) \\
Strict end-to-end success           & 69.2\% (18/26) \\
Correct abstention                  & 100.0\% (4/4) \\
Gold-source retrieval hit rate      & 100.0\% (17/17) \\
Gold-source citation hit rate       & 88.2\% (15/17) \\
Expected-source URL match (micro)   & 69.4\% (25/36) \\
Faithfulness: supported             & 72.7\% (16/22) \\
Faithfulness: partially supported   & 22.7\% (5/22) \\
Faithfulness: unsupported           & 4.5\% (1/22) \\
Forbidden-source citations          & 0 (0/5 applicable cases) \\
Mean E2E latency                    & 17.07~s \\
\bottomrule
\end{tabular}
\end{table}

\subsection{Discussion}

The results suggest that combining intent-aware retrieval with explicit evidence eligibility can improve bilingual consumer banking search over hybrid retrieval alone, most clearly at Top-1 (50.48\% to 79.05\%). The end-to-end run also shows promising source retrieval, citation, and abstention behavior in a fully local deployment. The evaluation is limited to a single-bank corpus, paraphrase-based retrieval queries, and a small end-to-end test set judged automatically; future work will use independently collected queries, human evaluation, and component-level ablations.

\section{Conclusion}
\label{conc}

This work presented an evidence-gated bilingual RAG framework for consumer banking search that combines retrieval-ready web-content refinement, hybrid dense--lexical retrieval, intent-aware evidence selection, and explicit answer eligibility within a local architecture. The proposed framework improved Top-1 hit rate from 50.48\% to 79.05\% and MRR@5 from 0.571 to 0.797 over the hybrid baseline on 210 URL-labeled queries, while a small end-to-end evaluation achieved 69.2\% strict success. These results are preliminary; future work will add independently collected queries and human evaluation.


\begin{thebibliography}{99}
\bibitem{scott2025} M. Scott, A. Acero, J. Apostolopoulos, L. Karam, M. Condry, R. Robson, and W. Talukdar, ``The 1st industry forum on large language models in consumer technology at IEEE ICCE 2025,'' \textit{IEEE Consum. Electron. Mag.}, vol. 14, pp. 8--11, 2025, doi: 10.1109/MCE.2025.3546425.
\bibitem{karpukhin2020} V. Karpukhin \textit{et al.}, ``Dense passage retrieval for open-domain question answering,'' in \textit{Proc. EMNLP}, 2020, pp. 6769--6781.
\bibitem{ma2021} X. Ma \textit{et al.}, ``A replication study of dense passage retriever,'' arXiv:2104.05740, 2021.
\bibitem{lewis2020} P. Lewis \textit{et al.}, ``Retrieval-augmented generation for knowledge-intensive NLP tasks,'' in \textit{Proc. NeurIPS}, 2020.
\bibitem{robertson2009} S. Robertson and H. Zaragoza, ``The probabilistic relevance framework: BM25 and beyond,'' \textit{Found. Trends Inf. Retr.}, vol. 3, no. 4, pp. 333--389, 2009.
\bibitem{nogueira2019} R. Nogueira and K. Cho, ``Passage re-ranking with BERT,'' arXiv:1901.04085, 2019.
\bibitem{cheng2026} Z. Cheng \textit{et al.}, ``Enhancing financial report question-answering: A retrieval-augmented generation system with reranking analysis,'' arXiv:2603.16877, 2026.
\bibitem{chen2024} J. Chen \textit{et al.}, ``BGE M3-Embedding: Multi-lingual, multi-functionality, multi-granularity text embeddings through self-knowledge distillation,'' arXiv:2402.03216, 2024.
\bibitem{chirkova2024} N. Chirkova \textit{et al.}, ``Retrieval-augmented generation in multilingual settings,'' arXiv:2407.01463, 2024.
\bibitem{elbeltagy2024} S. R. El-Beltagy and M. A. Abdallah, ``Exploring retrieval augmented generation in Arabic,'' arXiv:2408.07425, 2024.
\bibitem{hamed2025} I. Hamed \textit{et al.}, ``A survey of code-switched Arabic NLP: Progress, challenges, and future directions,'' arXiv:2501.13419, 2025.
\bibitem{obeid2020} O. Obeid \textit{et al.}, ``CAMeL tools: An open source Python toolkit for Arabic natural language processing,'' in \textit{Proc. LREC}, 2020, pp. 7022--7032.
\bibitem{asai2024} A. Asai, Z. Wu, Y. Wang, A. Sil, and H. Hajishirzi, ``Self-RAG: Learning to retrieve, generate, and critique through self-reflection,'' in \textit{Proc. ICLR}, 2024.
\bibitem{yan2024} S.-Q. Yan, J.-C. Gu, Y. Zhu, and Z.-H. Ling, ``Corrective retrieval augmented generation,'' arXiv:2401.15884, 2024.
\bibitem{kirichenko2025} P. Kirichenko \textit{et al.}, ``AbstentionBench: Reasoning LLMs fail on unanswerable questions,'' arXiv:2506.09038, 2025.
\bibitem{ji2023} Z. Ji \textit{et al.}, ``Survey of hallucination in natural language generation,'' \textit{ACM Comput. Surv.}, vol. 55, no. 12, pp. 1--38, 2023.
\bibitem{shang2024} Y. Shang, Z. Liu, J. Kang, M. S. Hossain, and Y. Wu, ``Adversarial attacks on vision-language model-empowered chatbots in consumer electronics,'' \textit{IEEE Trans. Consum. Electron.}, vol. 70, no. 3, pp. 6075--6083, 2024, doi: 10.1109/TCE.2024.3417688.
\bibitem{zhang2025} C. Zhang, G. Shan, B.-H. Roh, and J. Jiang, ``FM$^2$ Learning: LLM-based federated multi-task multi-domain learning for consumer electronics and IoT enhancement,'' \textit{IEEE Trans. Consum. Electron.}, 2025, doi: 10.1109/TCE.2025.3603592.
\bibitem{otoum2026} Y. Otoum, S. M. Danish, I. Ahmad, Y. Alkhrijah, and A. Asad, ``Efficient federated LLM framework for intelligent IoMT management,'' \textit{IEEE Trans. Consum. Electron.}, vol. 72, no. 3, pp. 5832--5847, 2026, doi: 10.1109/TCE.2026.3689383.
\bibitem{huang2026} B. Huang, X. Pan, H. Chai, F. Zhu, D. Liu, and X. Pan, ``High sample rate audio generation using neural large language models for mitigating device manufacturing defects,'' in \textit{Proc. IEEE Int. Conf. Consum. Electron. (ICCE)}, 2026, pp. 1--4, doi: 10.1109/ICCE67443.2026.11449771.
\bibitem{wang2025} Y. Wang, X. Li, H. Yu, F. Hu, G. Wang, and D. Lei, ``Continuous entity reasoning for multi-turn medical dialog generation,'' \textit{IEEE Trans. Consum. Electron.}, vol. 71, no. 4, pp. 10681--10694, 2025, doi: 10.1109/TCE.2025.3598217.
\bibitem{maiya2025} A. S. Maiya, ``OnPrem.LLM: A privacy-conscious document intelligence toolkit,'' arXiv:2505.07672, 2025.
\bibitem{cormack2009} G. V. Cormack, C. L. A. Clarke, and S. Buettcher, ``Reciprocal rank fusion outperforms Condorcet and individual rank learning methods,'' in \textit{Proc. SIGIR}, 2009, pp. 758--759.
\bibitem{bge_reranker_v2_m3} Beijing Academy of Artificial Intelligence, ``BAAI/bge-reranker-v2-m3 model card,'' Hugging Face. [Online]. Available: https://huggingface.co/BAAI/bge-reranker-v2-m3. Accessed: Oct. 10, 2026.
\bibitem{qwen2025} Qwen Team, ``Qwen3 technical report,'' arXiv:2505.09388, 2025.
\end{thebibliography}
\end{document}